\documentclass[final,5p,11pt,twocolumn,authoryear]{elsarticle}

\usepackage{graphicx,amsmath,url,amsfonts,booktabs,nicefrac,multirow,microtype,bm}

\def\rot#1{\rotatebox{90}{#1}}

\usepackage{hyperref,cleveref}
\hypersetup{colorlinks=true,linkcolor=blue,urlcolor=blue,citecolor=blue}

\journal{Journal of Medical Image Analysis}

\makeatletter
\def\ps@pprintTitle{%
 \let\@oddhead\@empty
 \let\@evenhead\@empty
 \def\@oddfoot{}%
 \let\@evenfoot\@oddfoot}
\makeatother

\begin{document}

\title{Training recurrent neural networks robust to incomplete data: application to Alzheimer's disease progression modeling}

\author[add1,add11,add2,add3]{Mostafa Mehdipour Ghazi\corref{cor1}}
\ead{mehdipour@biomediq.com}
\author[add1,add11,add2]{Mads Nielsen}
\author[add1,add11,add2]{Akshay Pai}
\author[add3,add4]{M. Jorge Cardoso}
\author[add3,add4]{Marc Modat}
\author[add3,add4]{S\'ebastien Ourselin}
\author[add1,add11,add2]{Lauge S{\o}rensen}
\author{for the Alzheimer's Disease Neuroimaging Initiative\corref{cor2}}

\cortext[cor1]{Corresponding Author.}
\cortext[cor2]{Data used in preparation of this article were obtained from the Alzheimer's Disease Neuroimaging Initiative (ADNI) database (adni.loni.usc.edu). As such, the investigators within the ADNI contributed to the design and implementation of ADNI and/or provided data but did not participate in analysis or writing of this report. A complete listing of ADNI investigators can be found at \url{http://adni.loni.usc.edu/wp-content/uploads/how_to_apply/ADNI_Acknowledgement_List.pdf}}
\address[add1]{Biomediq A/S, Copenhagen, DK}
\address[add11]{Cerebriu A/S, Copenhagen, DK}
\address[add2]{Department of Computer Science, University of Copenhagen, Copenhagen, DK}
\address[add3]{Centre for Medical Image Computing, University College London, London, UK}
\address[add4]{School of Biomedical Engineering and Imaging Sciences, King's College London, London, UK}

\begin{frontmatter}

\begin{abstract}
Disease progression modeling (DPM) using longitudinal data is a challenging machine learning task. Existing DPM algorithms neglect temporal dependencies among measurements, make parametric assumptions about biomarker trajectories, do not model multiple biomarkers jointly, and need an alignment of subjects' trajectories. In this paper, recurrent neural networks (RNNs) are utilized to address these issues. However, in many cases, longitudinal cohorts contain incomplete data, which hinders the application of standard RNNs and requires a pre-processing step such as imputation of the missing values. Instead, we propose a generalized training rule for the most widely used RNN architecture, long short-term memory (LSTM) networks, that can handle both missing predictor and target values. The proposed LSTM algorithm is applied to model the progression of Alzheimer's disease (AD) using six volumetric magnetic resonance imaging (MRI) biomarkers, i.e., volumes of ventricles, hippocampus, whole brain, fusiform, middle temporal gyrus, and entorhinal cortex, and it is compared to standard LSTM networks with data imputation and a parametric, regression-based DPM method. The results show that the proposed algorithm achieves a significantly lower mean absolute error (MAE) than the alternatives with $p < 0.05$ using Wilcoxon signed rank test in predicting values of almost all of the MRI biomarkers. Moreover, a linear discriminant analysis (LDA) classifier applied to the predicted biomarker values produces a significantly larger area under the receiver operating characteristic curve (AUC) of 0.90 vs. at most 0.84 with $p < 0.001$ using McNemar's test for clinical diagnosis of AD. Inspection of MAE curves as a function of the amount of missing data reveals that the proposed LSTM algorithm achieves the best performance up until more than 74\% missing values. Finally, it is illustrated how the method can successfully be applied to data with varying time intervals. This paper shows that built-in handling of missing values in training an LSTM network benefits the application of RNNs in neurodegenerative disease progression modeling in longitudinal cohorts.
\end{abstract}


\end{frontmatter}

{\let\thefootnote\relax\footnote{{This document is the accepted version of the manuscript published in Medical Image Analysis in Volume 53, Pages 39-46, with DOI: \url{https://doi.org/10.1016/j.media.2019.01.004}. \copyright 2019. This manuscript version is made available under the CC-BY-NC-ND 4.0 license http://creativecommons.org/licenses/by-nc-nd/4.0/.}}}

\section{Introduction}

Alzheimer's disease (AD) is a chronic neurodegenerative disorder that begins with memory loss and develops over time, causing issues in conversation, orientation, and control of bodily functions \citep{mckhann1984}. Early diagnosis of the disease is challenging and is usually made once cognitive impairment has already compromised daily living. Hence, developing robust, data-driven methods for disease progression modeling (DPM) utilizing longitudinal data is necessary to yield a complete perspective on the disease for better diagnosis, monitoring, and prognosis \citep{oxtoby2017}.

Existing longitudinal DPM methods model biomarkers as a function of disease progression using continuous curve fitting. In the AD progression modeling literature, a variety of regression-based methods have been proposed to fit logistic or polynomial functions to the longitudinal dynamic of each biomarker \citep{jedynak2012,fjell2013,oxtoby2014,donohue2014,yau2015,guerrero2016}. However, parametric assumptions on the biomarker trajectories not only limit the flexibility of such methods but also lead to the necessity of aligning subjects' trajectories. In addition, the existing approaches mostly rely on independent biomarker modeling, and none of them consider the temporal dependencies among measurements.

Recurrent neural networks (RNNs) are non-parametric sequence based learning methods that, by design, do not require alignment of subject trajectories. They offer continuous, joint modeling of longitudinal data while taking temporal dependencies among measurements into account \citep{pearlmutter1989}. Long short-term memory (LSTM) networks, the most widely used type of RNNs, developed to effectively capture long-term temporal dependencies by dealing with the exploding and vanishing gradient problem during backpropagation through time \citep{hochreiter1997,gers1999,gers2001}. They employ a memory cell with nonlinear reset units -- so called constant error carousels (CECs) -- and learn to store history for either long or short time periods. Since their introduction, a variety of LSTM networks have been developed for different time-series applications \citep{greff2017}. The vanilla LSTM that utilizes three reset gates with full gate recurrence is the most commonly used LSTM architecture. It applies the backpropagation through time algorithm using full gradients to train the network and can include biases and cell-to-gates (peephole) connections.

However, since longitudinal cohorts often contain missing biomarker values due to, for instance, dropped out patients, unsuccessful measurements, or different assessment patterns used for different subject groups -- as seen in the Alzheimer's Disease Neuroimaging Initiative (ADNI) \citep{Petersen2010}, standard RNNs inclunding LSTMs cannot be directly applied. Pre-processing methods such as data imputation and interpolation are the most common approaches to handling missing data in RNNs. These two-step procedures decouple missing data handling and network training, resulting in a sub-optimal performance that is heavily influenced by the choice of data pre-processing method \citep{lipton2016}. Although RNNs themselves have been used for estimating missing data \citep{parveen2002,yoon2017}, the lack of methods to inherently handle incomplete data in RNNs is evident \citep{che2016}. Other approaches update the architecture to learn or encode the missing data patterns \citep{che2016,lipton2016}. These methods are typically biased towards specific cohort or demographic circumstances correlated with the learned missing data patterns and introduce additional parameters in the network which increases the complexity of the network.

In this paper, we propose a generalized method for training LSTM networks that can handle missing values in both input and target. This is achieved by applying the batch gradient descent algorithm in combination with the loss function and its gradients normalized by the number of missing values in input and target. Our goal is different than the approaches that encode the missing values' patterns \citep{che2016,lipton2016}; we want to train RNNs robust to missing values to more faithfully capture the true underlying signal and to make the learned model generalizable across cohorts. The proposed LSTM algorithm is applied to AD progression modeling in the ADNI cohort \citep{Petersen2010} based on volumetric magnetic resonance imaging (MRI) biomarkers, and the estimated biomarker values are used to predict the clinical status of subjects. MRI is known to be the best noninvasive way to examine changes in the brain in vivo during the course of AD \citep{biagioni2011,wu2011}, and volumetric analysis is a widely used ROI-based method to estimate brain atrophy.

\begin{figure*}[t]
\centering
\includegraphics[scale=0.59]{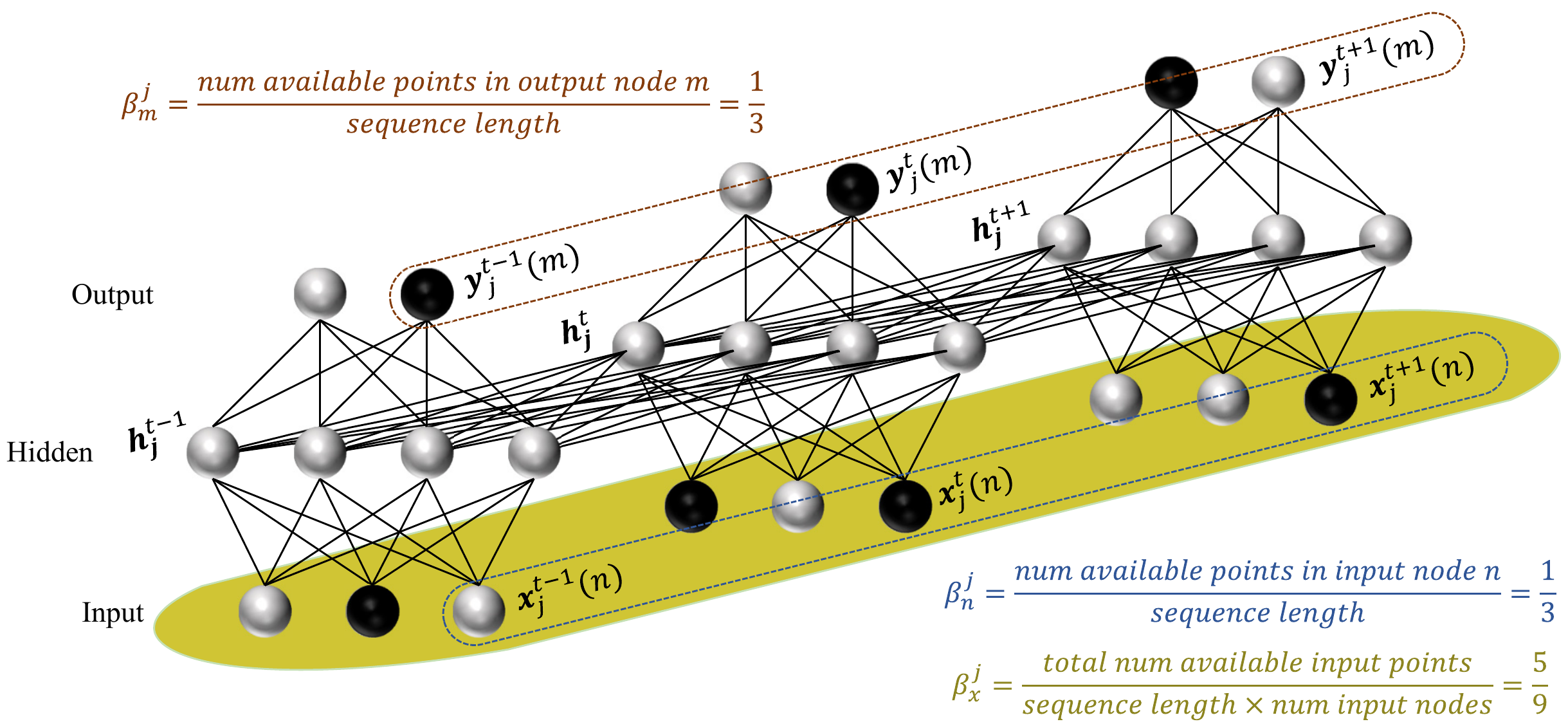}
\caption{Illustration of how the normalization factors are related to the input and output of an unfolded RNN. Assume an RNN with three consecutive time points $\{t-1, t, t+1\}$, three input nodes, four hidden nodes, and two output nodes. Missing data for an instance observation $j$ is illustrated as black nodes. We wish to weight the loss function and its gradients according to the number of available points in the input and output nodes. In this specific example, subject $j$ has only one measurement available for its $n$-th input node and the same many for its $m$-th output node. Hence, the loss function and its gradients are weighted by 1/3. Moreover, since there is a total of five measurements available in the input layer, the loss function is weighted by 5/9. The later weighting factor is to ensure that the loss function takes the number of available points in the input layer into account.}
\label{figure2}
\end{figure*}

The main contribution is three-fold and can be summarized as follows:
\begin{itemize}
\item First, a generalized formulation of backpropagation through time for LSTM networks is proposed to handle incomplete data, and it is shown that such built-in handling of missing values provides a better modeling and prediction performance compared to using data imputation with standard LSTM networks. 
\item Second, temporal dependencies among measurements in the ADNI data are modeled using the proposed LSTM network via sequence-to-sequence learning. To the best of our knowledge, this is the first time such multi-dimensional sequence learning methods are applied to neurodegenerative DPM. 
\item Third, an end-to-end approach, without need for trajectory alignment, is proposed for modeling the longitudinal dynamics of imaging biomarkers and for clinical status prediction. This is a practical way of implementing a robust DPM for both research and clinical applications.
\end{itemize}

A preliminary version of this work appeared in proceedings of the International Conference on Medical Imaging with Deep Learning \citep{Ghazi2018}. The present study contains a more detailed presentation and additional experiments to investigate statistical significance, robustness as a function of amount of missing data, and situations with varying time steps.

\section{Proposed LSTM algorithm}

The main goal of this study is to minimize the influence of missing values on the learned LSTM network parameters. This is achieved by using the batch gradient descend method in combination with the backpropagation through time algorithm modified to take into account missing values in the input and target vectors. More specifically, the algorithm sets input missing values to zero, backpropagates zero errors corresponding to the target missing points, and uses an L2-norm loss function with residuals weighted according to the number of available time points per target biomarker node ($\beta_m^j$) and according to the total number of available input values for all visits of all biomarkers ($\beta_x^j$). In addition, it normalizes input weight gradients of the loss function according to the number of available time points per input biomarker node ($\beta_n^j$). Figure \ref{figure2} provides an illustration of how the normalization factors are related to the input and output of an unfolded RNN. Note that the use of batch gradient descend ensures the availability of at least one data point per biomarker that can proportionally contribute in the weight update rule.

\subsection{The basic LSTM architecture}

\begin{figure}
\centering
\includegraphics[scale=0.6]{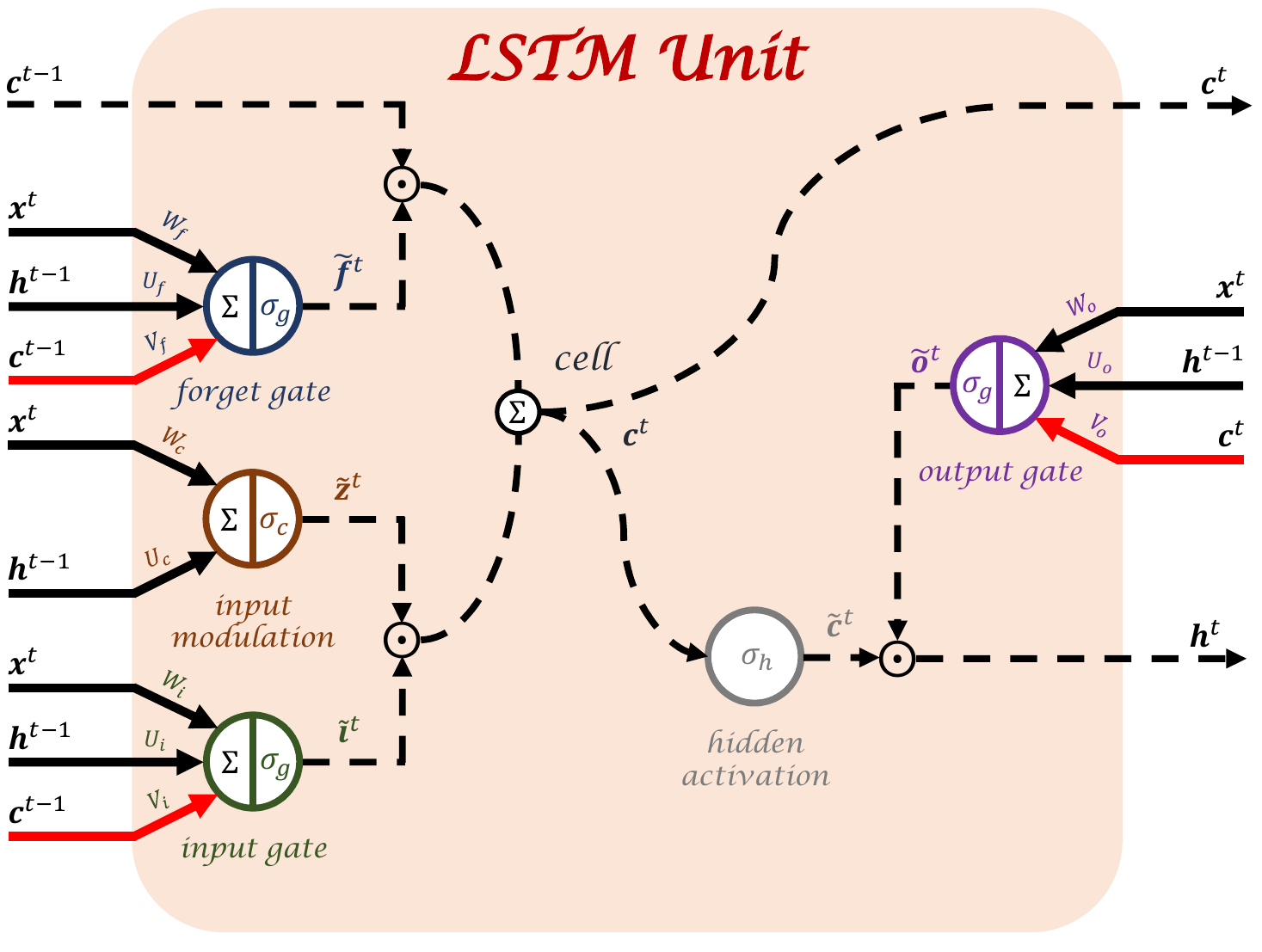}
\caption{An illustration of a vanilla LSTM unit with peephole connections in red. The solid and dashed lines show weighted and unweighted connections, respectively.}
\label{figure1}
\end{figure}

Figure \ref{figure1} shows a typical schematic of a vanilla LSTM architecture. As can be seen, the topology includes a memory cell, an input modulation gate, and three nonlinear reset gates, namely input gate, forget gate, and output gate, each of which accepting current and recurrent inputs. The memory cell learns to maintain its state over time while the multiplicative gates learn to open and close access to the constant error/information flow, to prevent exploding or vanishing gradients. The input gate protects the memory contents from perturbation by irrelevant inputs, and the output gate protects other units from perturbation by currently irrelevant memory contents. The forget gate deals with continual or very long input sequences, and finally, peephole connections allow the gates to access the CEC of the same cell state.

\subsection{Feedforward in LSTM networks}

Assume $\bm{x}_j^t\in\mathbb{R}^{N\times1}$ is the $j$-th observation of an $N$-dimensional input vector at current time $t$. If $M$ is the number of output units, feedforward calculations of the LSTM network under study can be summarized as
\begin{flalign*}
\bm{f}_j^t &= W_f\bm{x}_j^t+U_f\bm{h}_j^{t-1}+\bm{V}_f\odot \bm{c}_j^{t-1}+\bm{b}_f \,, &\\
\bm{\tilde{f}}_j^t &= \sigma_g(\bm{f}_j^t ) \,, &\\
\bm{i}_j^t &= W_i\bm{x}_j^t+U_i\bm{h}_j^{t-1}+\bm{V}_i\odot \bm{c}_j^{t-1}+\bm{b}_i \,, &\\
\bm{\tilde{i}}_j^t &= \sigma_g(\bm{i}_j^t ) \,, &\\
\bm{z}_j^t &= W_c\bm{x}_j^t+U_c\bm{h}_j^{t-1}+\bm{b}_c \,, &\\
\bm{\tilde{z}}_j^t &= \sigma_c(\bm{z}_j^t) \,, &\\
\bm{c}_j^t &= \bm{\tilde{f}}_j^t\odot \bm{c}_j^{t-1}+\bm{\tilde{i}}_j^t\odot\bm{\tilde{z}}_j^t \,, &\\
\bm{\tilde{c}}_j^t &= \sigma_h(\bm{c}_j^t) \,, &\\
\bm{o}_j^t &= W_o\bm{x}_j^t+U_o\bm{h}_j^{t-1}+\bm{V_o}\odot \bm{c}_j^t+\bm{b}_o \,, &\\
\bm{\tilde{o}}_j^t &= \sigma_g(\bm{o}_j^t ) \,, &\\
\bm{h}_j^t &= \bm{\tilde{o}}_j^t\odot \bm{\tilde{c}}_j^t \,,
\end{flalign*}

\noindent where $\{\bm{f}_j^t,\bm{i}_j^t,\bm{z}_j^t,\bm{c}_j^t,\bm{o}_j^t,\bm{h}_j^t\}\in\mathbb{R}^{M\times1}$ and $\{\bm{\tilde{f}}_j^t,\bm{\tilde{i}}_j^t,\bm{\tilde{z}}_j^t,\bm{\tilde{c}}_j^t,\bm{\tilde{o}}_j^t\}\in\mathbb{R}^{M\times1}$ are $j$-th observation of forget gate, input gate, modulation gate, cell state, output gate, and hidden output at time $t$ before and after activation, respectively. Moreover, $\{W_f,W_i,W_o,W_c\}\in\mathbb{R}^{M\times N}$ and $\{U_f,U_i,U_o,U_c\}\in\mathbb{R}^{M\times M}$ are sets of connecting weights from current and recurrent inputs to the gates and cell, respectively, $\{\bm{V}_f,\bm{V}_i,\bm{V}_o\}\in\mathbb{R}^{M\times1}$ is the set of peephole connections from the cell to the gates, $\{\bm{b}_f,\bm{b}_i,\bm{b}_o,\bm{b}_c\}\in\mathbb{R}^{M\times1}$ represents corresponding biases of neurons, and $\odot$ denotes element-wise multiplication. Finally, $\sigma_g$, $\sigma_c$, and $\sigma_h$ are nonlinear activation functions assigned for the gates, input modulation, and hidden output, respectively. Logistic sigmoid functions are applied to the gates with range $\left[0, 1\right]$ while hyperbolic tangent functions are applied to modulate both cell input and hidden output with range $\left[-1, 1\right]$. Hence, the measurements need to be in the same range $\left[-1, 1\right]$.

\subsection{Robust backpropagation through time}

Let $\bm{\mathcal{L}}\in\mathbb{R}^{M\times1}$ be the loss function defined based on the actual target $\bm{s}$ and network output $\bm{y}$. Here, we consider one layer of LSTM units for sequence learning which means that the network output is the hidden output. The main idea is to calculate the partial derivatives of the normalized loss function ($\delta$) with respect to the weights using the chain rule.
\begin{flalign*}
&\bm{\mathcal{L}}(m) = \frac{1}{2JT} \sum_{j,t}\frac{1}{\beta_x^j \beta_m^j}(\bm{y}_j^t(m)-\bm{s}_j^t(m))^2 \,, &\\
&\delta\bm{y}_j^t(m) = \frac{1}{JT} \Big[\frac{1}{\beta_x^j\beta_m^j}(\bm{y}_j^t(m)-\bm{s}_j^t(m))\Big] \,,
\end{flalign*}
where $\beta_x^j=\frac{|\bm{x}_j|}{TN}$ and $\beta_m^j=\frac{|\bm{y}_j(m)|}{T}$ are normalization factors to handle missing values of the $j$-th observation with batch size $J$ and sequence length $T$. Also, $|\bm{x}_j|$ and $|\bm{y}_j(m)|$ denote the total number of available input values and the number of available target time points in the $m$-th node, respectively. The backpropagation calculations through time using full gradients can be obtained as
\begin{flalign*}
\nonumber \delta\bm{h}_j^t &= U_f^T\delta\bm{f}_j^{t+1}+U_i^T\delta\bm{i}_j^{t+1}+U_c^T \delta\bm{z}_j^{t+1}+U_o^T \delta\bm{o}_j^{t+1} &\\
&+\delta\bm{y}_j^t \,, &\\
\delta\bm{\tilde{o}}_j^t &= \delta\bm{h}_j^t\odot\bm{\tilde{c}}_j^t \,, &\\
\delta\bm{o}_j^t &= \delta\bm{\tilde{o}}_j^t\odot\sigma'_g(\bm{o}_j^t) \,, &\\
\delta\bm{\tilde{c}}_j^t &= \delta\bm{h}_j^t \odot\bm{\tilde{o}}_j^t \,, &\\
\nonumber \delta\bm{c}_j^t &= \bm{V}_f\odot\delta\bm{f}_j^{t+1}+\bm{V}_i\odot\delta\bm{i}_j^{t+1}+\bm{V}_o\odot\delta\bm{o}_j^t &\\
&+\delta\bm{\tilde{c}}_j^t\odot\sigma'_h(\bm{c}_j^t)+\delta\bm{c}_j^{t+1}\odot\bm{\tilde{f}}_j^{t+1} \,, &\\
\delta\bm{\tilde{z}}_j^t &= \delta\bm{c}_j^t\odot\bm{\tilde{i}}_j^t \,, &\\
\delta\bm{z}_j^t &= \delta\bm{\tilde{z}}_j^t\odot\sigma'_c(\bm{z}_j^t) \,, &\\
\delta\bm{\tilde{i}}_j^t &= \delta\bm{c}_j^t\odot\bm{\tilde{z}}_j^t \,, &\\
\delta\bm{i}_j^t &= \delta\bm{\tilde{i}}_j^t\odot\sigma'_g(\bm{i}_j^t) \,, &\\
\delta\bm{\tilde{f}}_j^t &= \delta\bm{c}_j^t\odot\bm{c}_j^{t-1} \,, &\\
\delta\bm{f}_j^t &= \delta\bm{\tilde{f}}_j^t\odot\sigma'_g(\bm{f}_j^t) \,, &\\
\delta\bm{x}_j^t &= W_f^T\delta\bm{f}_j^t+W_i^T\delta\bm{i}_j^t+W_c^T\delta\bm{z}_j^t+W_o^T\delta\bm{o}_j^t \,,
\end{flalign*}

Finally, if $\theta\in\{f,i,z,o\}$ and $\phi\in\{f,i\}$, the gradients of the loss function with respect to the weights are calculated as
\begin{flalign*}
&\delta W_\theta(n) = \sum_{j=1}^J\frac{1}{\beta_n^j}\delta\bm{\theta}_j^{\{0\rightarrow T\}} \bm{x}_j^{\{0\rightarrow T\}}(n) \,, &\\
&\delta U_\theta = \sum_{j=1}^J\delta\bm{\theta}_j^{\{1\rightarrow T\}} \bm{h}_j^{\{0\rightarrow T-1\}} \,, &\\
&\delta \bm{V}_\phi = \sum_{j=1}^J\sum_{t=0}^{T-1}\delta\bm{\phi}_j^{t+1}\odot\bm{c}_j^t \,, &\\
&\delta \bm{V}_o = \sum_{j=1}^J\sum_{t=0}^T\delta\bm{o}_j^t\odot\bm{c}_j^t \,, &\\
&\delta \bm{b}_\theta = \sum_{j=1}^J\sum_{t=0}^T\delta\bm{\theta}_j^t \,,
\end{flalign*}

\noindent where $\beta_n^j=\frac{|\bm{x}_j(n)|}{T}$ is the normalization factor handling missing input values and $|\bm{x}_j(n)|$ is the number of available time points in the input's $n$-th node. Here, we use a fixed sequence length of $T$ to proportionally consider subjects based on their available visits. However, the robust backpropagation algorithm can easily be generalized for a dynamic sequence length.

\subsection{Momentum batch gradient descent}

As an efficient iterative algorithm, momentum batch gradient descent is applied to find the local minimum of the loss function calculated over a batch while speeding up the convergence. The update rule using L2 regularization can be written as
\begin{flalign*}
&\vartheta^{new} = \mu\vartheta^{old}-\alpha(\delta\omega+\gamma\omega^{old}) \,, &\\
&\omega^{new} = \omega^{old}+\vartheta^{new} \,,
\end{flalign*}

\noindent where $\vartheta$ is the weight update initialized to zero, $\omega$ is the to-be-updated weight array, $\delta\omega$ is the gradient of the loss function with respect to $\omega$, and $\alpha$, $\gamma$, and $\mu$ are the learning rate, weight decay or regularization factor, and momentum weight, respectively.

\section{Experiments}

\subsection{Data}

Data used in the preparation of this article is obtained from the ADNI database. The ADNI was launched in 2003 as a public-private partnership, led by principal investigator Michael W. Weiner, MD. The primary goal of ADNI has been to test whether serial magnetic resonance imaging, positron emission tomography, other biological markers, and clinical and neuropsychological assessment can be combined to measure the progression of mild cognitive impairment and early Alzheimer's disease. To be more specific, we use The Alzheimer's Disease Prediction Of Longitudinal Evolution (TADPOLE) challenge dataset \citep{marinescu2018} which is composed of data from the three ADNI phases ADNI 1, ADNI GO, and ADNI 2. This includes roughly 1,500 biomarkers acquired from 1,737 subjects (957 males and 780 females) during 12,741 visits at 22 distinct time points between 2003 and 2017. Table \ref{table1} summarizes statistics of the demographics in the TADPOLE dataset. Note that the subjects include missing values and clinical status during their visits.

\begin{table*}[t]
\centering
\normalsize
\caption{Demographics of the TADPOLE dataset.}
\label{table1}
\renewcommand{\arraystretch}{1.15}
\centering
\begin{tabular}{lcccccc}
\toprule
 & \multicolumn{2}{c}{Number of visits} & \multicolumn{2}{c}{Age, year (mean$\pm$SD)} & \multicolumn{2}{c}{Education, year (mean$\pm$SD)} \\
 & male & female & male & female & male & female \\ \hline
\midrule
CN & 1,356 & 1,389 & 76.67$\pm$6.44 & 75.85$\pm$6.28 & 17.06$\pm$2.51 & 15.74$\pm$2.71 \\
MCI & 2,454 & 1,604 & 75.59$\pm$7.47 & 73.87$\pm$8.09 & 16.22$\pm$2.85 & 15.45$\pm$2.76 \\
AD & 1,208 & 900 & 77.22$\pm$7.11 & 75.45$\pm$7.92 & 15.85$\pm$3.03 & 14.35$\pm$2.73 \\
All (labeled \& unlabeled) & \multicolumn{2}{c}{12,741} & \multicolumn{2}{c}{76.00$\pm$7.38} & \multicolumn{2}{c}{15.91$\pm$2.86} \\
\bottomrule
\end{tabular}
\end{table*}

In this work, we have merged existing groups labeled as cognitively normal (CN), significant memory concern (SMC), and normal (NL) under CN, mild cognitive impairment (MCI), early MCI (EMCI), and late MCI (LMCI) under MCI, and Alzheimer's disease (AD) and dementia under AD. Moreover, groups with labels converting from one status to another, e.g. MCI-to-AD, belong to the next status (AD in this example).

MRI biomarkers are used for AD progression modeling. This includes T1-weighted brain MRI volumes of ventricles, hippocampus, whole brain, fusiform, middle temporal gyrus, and entorhinal cortex. We normalize the MRI measurements by the corresponding intracranial volume (ICV). Next, we filter within-class outliers of each biomarker -- across all subjects and their visits -- by assuming them as missing values and normalize the measurements by scaling them linearly to $\left[-1, 1\right]$. Out of 22 visits, we initially select 11 regular visits with a fixed interval of one year including baseline. Finally, subjects with less than three distinct visits for any biomarker are removed to obtain 742 subjects. This is to ensure that at least two visits are available per biomarker for performing sequence learning through the feedforward step and an additional visit for backpropagation.

For evaluation purpose, we partition the entire dataset to three non-overlapping subsets for training, validation, and testing. To achieve this, we randomly select 10\% of the within-class subjects for validation and the same for testing. More specifically, we randomly pick subjects based on their baseline labels while ensuring that subjects with few and large number of visits are included in each subset. This process results in 592, 76, and 74 subjects for training, validations, and testing, respectively. Details on the amount of available visits in the obtained evaluation subsets are shown in Table \ref{table2}. As can be deduced from the table, 63\% of the obtained data is missing.

\begin{table*}[t]
\centering
\scriptsize
\caption{Number of visits in the evaluation subsets across all subjects. Note that the complete dataset should have contained $742\times11=8,162$ visits per biomarker where the maximum number of visits per subject is 11. The number of visits per subject per diagnostic group is left blank as subjects can convert from one group to another in the course of AD.}
\label{table2}
\renewcommand{\arraystretch}{1.4}
\centering
\begin{tabular}{llcccccc}
\toprule
 & & Number of visits across subjects & \multicolumn{5}{c}{Number of visits per subject (mean$\pm$SD $\sim$ [min, max])} \\
 & & train / validation / test & train & / & validation & / & test \\ \hline
\midrule
& CN & 1,192 / 136 / 149 & & & & & \\
& MCI & 1,389 / 198 / 180 & & & & & \\
& AD & 606 / 84 / 92 & & & & & \\
\rot{\rlap{Clinical labels}} & All (labeled \& unlabeled) & 3,270 / 428 / 434 & 5.52$\pm$2.32 $\sim$ [3, 11] & / & 5.63$\pm$2.39 $\sim$ [3, 11] & / & 5.86$\pm$2.51 $\sim$ [3, 11] \\ \hline
& Ventricles & 2,481 / 328 / 318 & 4.19$\pm$1.47 $\sim$ [3, 10] & / & 4.32$\pm$1.46 $\sim$ [3, 8] & / & 4.30$\pm$1.58 $\sim$ [3, 9] \\
& Hippocampus & 2,381 / 311 / 312 & 4.02$\pm$1.31 $\sim$ [3, 10] & / & 4.09$\pm$1.29 $\sim$ [3, 8] & / & 4.22$\pm$1.49 $\sim$ [3, 7] \\
& Whole brain & 2,513 / 328 / 322 & 4.24$\pm$1.49 $\sim$ [3, 10] & / & 4.32$\pm$1.46 $\sim$ [3, 8] & / & 4.35$\pm$1.57 $\sim$ [3, 9] \\
& Entorhinal cortex & 2,351 / 310 / 309 & 3.97$\pm$1.29 $\sim$ [3, 10] & / & 4.08$\pm$1.34 $\sim$ [3, 8] & / & 4.18$\pm$1.46 $\sim$ [3, 7] \\
& Fusiform & 2,351 / 310 / 309 & 3.97$\pm$1.29 $\sim$ [3, 10] & / & 4.08$\pm$1.34 $\sim$ [3, 8] & / & 4.18$\pm$1.46 $\sim$ [3, 7] \\
\rot{\rlap{~~~MRI biomarkers}} & Middle temporal gyrus & 2,351 / 309 / 309 & 3.97$\pm$1.29 $\sim$ [3, 10] & / & 4.07$\pm$1.35 $\sim$ [3, 8] & / & 4.18$\pm$1.46 $\sim$ [3, 7] \\
\bottomrule
\end{tabular}
\end{table*}

\subsection{Evaluation metrics and statistical tests}

Mean absolute error (MAE) and multi-class area under the receiver operating characteristic (ROC) curve (AUC) are used to assess the performance of modeling and classification, respectively. MAE measures accuracy of continuous prediction per biomarker by computing the absolute difference between actual and estimated values as follows
\begin{flalign*}
&\mathrm{MAE} = \frac{1}{\mathcal{I}} \sum_{j,t}|\bm{y}_j^t-\bm{s}_j^t| \,, &
\end{flalign*}

\noindent where $\bm{s}_j^t$ and $\bm{y}_j^t$ are the ground-truth and estimated values of the specific biomarker for the $j$-th subject at the $t$-th visit, respectively, and $\mathcal{I}$ is the number of available points in the target array $\bm{s}$. 

Multi-class AUC \citep{hand2001} is a measure to examine the diagnostic performance in a multi-class test set using ROC analysis. It is calculated using the posterior probabilities as follows
\begin{flalign*}
\nonumber \mathrm{AUC} &= \frac{1}{(n_c(n_c-1))} \sum_{i=1}^{n_c-1}\sum_{k=i+1}^{n_c}\frac{1}{n_i n_k} &\\
& \times \Big[\mathrm{SR}_i - \frac{n_i(n_i+1)}{2}+\mathrm{SR}_k-\frac{n_k(n_k+1)}{2}\Big] \,, &
\end{flalign*}

\noindent where $n_c$ is the number of distinct classes, $n_i$ denotes the number of available points belonging to the $i$-th class, and $\mathrm{SR}_i$ is the sum of the ranks of posteriors $p(c_i|\bm{s}_i)$ after sorting all concatenated posteriors $\{p(c_i|\bm{s}_i),p(c_i|\bm{s}_k)\}$ in an ascending order, where $\bm{s}_i$ and $\bm{s}_k$ are vectors of scores belonging to the true classes $c_i$ and $c_k$, respectively.

The modeling performance is statistically assessed for different methods using the paired, two-sided Wilcoxon signed rank test \citep{wilcoxon1945} applied to the obtained absolute errors. Also, classification performance is analyzed using McNemar's test \citep{mcnemar1947} applied to the hard classification results (clinical status) obtained from a linear discriminant analysis (LDA) classifier with predicted MRI measurements as input.

\subsection{Experimental setup}

The following methods are evaluated in our conducted experiments:
\begin{itemize}
\item LSTM-Robust: an LSTM network trained based on the proposed robust backpropagation through time algorithm by setting input missing values to zero and backpropagating zero errors corresponding to the target missing points while training.
\item LSTM-Mean: an LSTM network trained using the standard backpropagation through time algorithm with missing values imputed based on mean imputation method prior to training \citep{che2016}.
\item LSTM-Forward: an LSTM network trained using the standard backpropagation through time algorithm with missing values imputed based on forward imputation method prior to training \citep{lipton2016}.
\item Regression-Based: a parametric, regression-based method \citep{jedynak2012} that automatically handles missing values. The parameters of the algorithm are initially estimated using linear regression in 15 iterations and are optimized using sigmoidal functions in 35 additional iterations where all parameters converge.
\end{itemize}

All the methods are developed in MATLAB R2017b and run on a 2.80 GHz CPU with 16 GB RAM. We initialize the LSTM networks' weights by generating uniformly distributed random values in range $[-0.05, 0.05]$ and set the weights' updates and weights' gradients to zero. The batch size is set to the number of available training subjects, and the first ten visits are used to estimate the second to eleventh visits per subject for evaluation purpose. It should be noted that when data imputation is applied, the robust backpropagation formulas simply generalize to the ones for the standard LSTM network.

We utilize the validation set to tune all the networks' optimization parameters, each time by adjusting one of the parameters while keeping the rest at fixed values to achieve the lowest average MAE. Peephole connections are used in the networks since they tend to improve the performance \citep{greff2017}. Based on these strategies, the optimal parameters are obtained as $\alpha = 0.1$, $\mu = 0.9$, and $\gamma = 0.0001$ with 1,000 epochs. The corresponding MAEs for the validation set are also calculated as 0.00296, 0.00025, 0.01494, 0.00024, 0.00076, and 0.00097, for ventricles, hippocampus, whole brain, entorhinal cortex, fusiform, and middle temporal gyrus, respectively. It takes about 340 seconds to train the network and 0.025 seconds to estimate all the validation measurements. It is worthwhile mentioning that all the estimated measurements are linearly scaled from $\left[-1, 1\right]$ to the original range of biomarkers using the original minimum and maximum values while calculating MAEs.

\section{Results and discussion}

After successfully training the LSTM networks and the regression-based method for DPM, they are all evaluated using the test set.

\subsection{Biomarker modeling}

Table \ref{table3} compares the test MRI biomarker modeling performance (MAE) using aforementioned methods. Even though the performance is reported per biomarker, the models are jointly fitted to all biomarkers. As it can be deduced from Table \ref{table3}, LSTM-Robust significantly outperforms the other methods in all MRI biomarkers except for whole brain where the regression-based approach performs significantly better and for middel temporal gyrus where there is no difference between the proposed method and LSTM-Forward.

\begin{table*}[t]
\centering
\normalsize
\caption{Test MRI biomarker modeling performance (MAE) for yearly predictions. The proposed method is compared with the alternatives using a paired, two-sided Wilcoxon signed rank test, and this is reported in superscript as LSTM-Robust vs. LSTM-Mean/LSTM-Robust vs. LSTM-Forward/LSTM-Robust vs. Regression-Based. $\dagger:$ not significantly different, $\star: p < 0.05$, $\star\star: p < 0.01$, $\star\star\star: p < 0.001$.}
\label{table3}
\renewcommand{\arraystretch}{1.15}
\begin{tabular}{llccc}
\toprule
 & LSTM-Robust & LSTM-Mean & LSTM-Forward & Regression-Based \\
 & & \citep{che2016} & \citep{lipton2016} & \citep{jedynak2012} \\ \hline
\midrule
Ventricles & $0.00307^{\star\star\star/\star\star\star/\star\star\star}$ & $0.00620$ & $0.00472$ & $0.00807$ \\
Hippocampus & $0.00023^{\star\star\star/\star\star/\star\star\star}$ & $0.00051$ & $0.00034$ & $0.00051$ \\
Whole brain & $0.01330^{\star\star\star/\star\star/\star\star\star}$ & $0.02375$ & $0.01639$ & $0.00551$ \\
Entorhinal cortex & $0.00021^{\star\star\star/\star/\star\star\star}$ & $0.00030$ & $0.00025$ & $0.00035$ \\ 
Fusiform & $0.00068^{\star\star\star/\star\star\star/\star\star\star}$ & $0.00130$ & $0.00100$ & $0.00090$ \\
Middle temporal gyrus & $0.00087^{\star\star\star/\dagger/\star}$ & $0.00126$ & $0.00118$ & $0.00111$ \\
\bottomrule
\end{tabular}
\end{table*}

\subsection{Predicting clinical status}

To assess the ability of the estimated measurements in predicting the clinical status, we train an LDA classifier using the estimated training measurements and apply it to the estimated test data to compute the posterior probabilities. The obtained scores are then used to calculate diagnostic AUCs. The diagnostic prediction results for the test set are shown in Table \ref{table4}. As can be seen, LSTM-Robust outperforms all other methods in predicting clinical status of subjects per visit with a multi-class AUC of 0.76, which reveals the effect of modeling on classification performance. One could of course use other classifiers or train the LSTM network directly for classification based on sequence-to-label learning to potentially improve the diagnostic AUCs. However, the focus of this work is on DPM based on sequence-to-sequence learning. In addition, sequence-to-label learning would only be able to utilize the part of the training data which has available clinical status.

\begin{table*}[t]
\centering
\normalsize
\caption{Test diagnostic performance (AUC) of the estimated MRI biomarker values using an LDA classifier. The proposed method is compared with the alternatives using McNemar's test, and this is reported in superscript as LSTM-Robust vs. LSTM-Mean/LSTM-Robust vs. LSTM-Forward/LSTM-Robust vs. Regression-Based. $\dagger:$ not significantly different, $\star: p < 0.05$, $\star\star: p < 0.01$, $\star\star\star: p < 0.001$.}
\label{table4}
\renewcommand{\arraystretch}{1.15}
\begin{tabular}{llccc}
\toprule
 & LSTM-Robust & LSTM-Mean & LSTM-Forward & Regression-Based \\
 & & \citep{che2016} & \citep{lipton2016} & \citep{jedynak2012} \\ \hline
\midrule
CN vs. MCI & $0.5914^{\dagger/\dagger/\dagger}$ & 0.5838 & 0.5800 & 0.5468 \\
CN vs. AD & $0.9029^{\star\star\star/\star\star\star/\star\star\star}$ & 0.8404 & 0.8150 & 0.7826 \\
MCI vs. AD & $0.7844^{\dagger/\dagger/\dagger}$ & 0.6936 & 0.6890 & 0.7330 \\
CN vs. MCI vs. AD & $0.7596^{\dagger/\star/\star}$ & 0.7059 & 0.6947 & 0.6875 \\
\bottomrule
\end{tabular}
\end{table*}

The multi-class AUC of 0.76 obtained using predicted measurements from the proposed approach is within the top-five AUCs of the state-of-the-art, cross-sectional MRI-based classification results of the recent challenge on Computer-Aided Diagnosis of Dementia (CADDementia) \citep{bron2015} that ranged from 0.75 to 0.79. It should, however, be noted that there are important differences between this study and the CADDementia challenge. Firstly, this work has the advantage of training and testing data from the same cohort whereas CADDementia algorithms were applied to classify data from independent cohorts. Secondly, the top performing CADDementia algorithms incorporated different types of MRI biomarkers besides volumetry. Thirdly, this work predicts the input features to the classifier based on historical longitudinal data.

\subsection{Robustness as a function of amount of missing data}

\begin{figure}[t]
\centering
\includegraphics[scale=0.565]{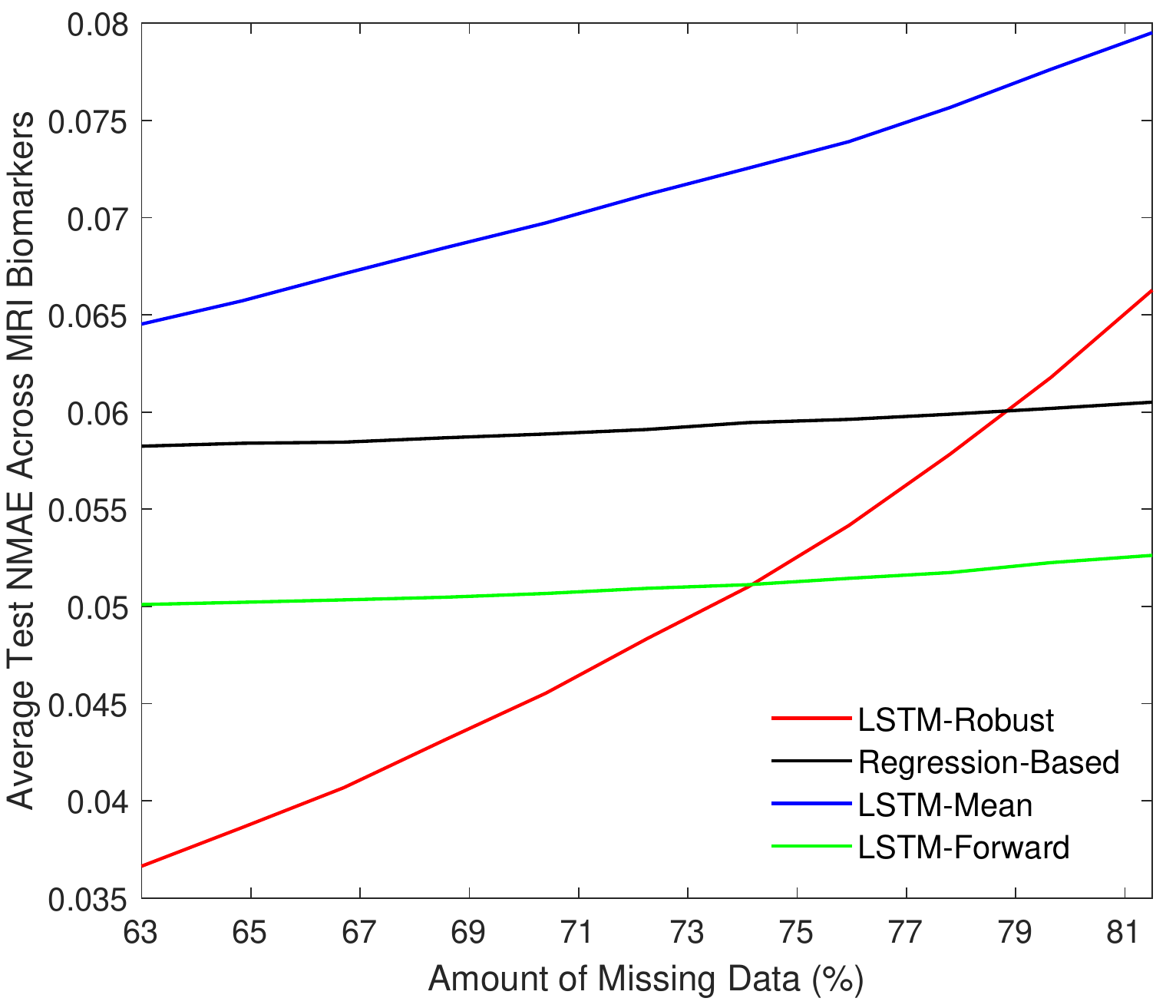}
\caption{Modeling performance of MRI biomarkers for various amounts of missing values.}
\label{figure3}
\end{figure}

To evaluate the modeling robustness of the proposed method compared to the alternatives for different amounts of missing data, we construct subsamples of the training dataset by randomly removing up to 50\% of the actual data per biomarker and train the methods on the smaller datasets. Figure \ref{figure3} illustrates the modeling performance of the different methods on various amounts of missing measurements, from 0\% to 50\%. It is important to note that the training data already includes a large number of missing values at missing rate of 0\% -- i.e. 63\% of actual data as seen on Table \ref{table2}. For better comparison, we take the average of MAEs normalized by the range of corresponding biomarkers to obtain a single curve per method. As can be seen, the result of the proposed method is superior to those of the benchmarks up until missing around 74\% of the data. For higher rates of missing data, basic LSTM with forward imputation outperforms all other methods. One reason for why LSTM with forward imputation is robust to the higher rates of missing data could be due to the fact that it replaces the missing values placed at the beginning of a sequence with the whole training data median.

\begin{table*}[t]
\centering
\normalsize
\caption{Test MRI biomarker modeling performance (MAE) for half-yearly predictions.}
\label{table6}
\renewcommand{\arraystretch}{1.15}
\begin{tabular}{lcccc}
\toprule
 & LSTM-Robust & LSTM-Mean & LSTM-Forward & Regression-Based \\
 & & \citep{che2016} & \citep{lipton2016} & \citep{jedynak2012} \\ \hline
\midrule
Ventricles & $0.00272$ & $0.00973$ & $0.01030$ & $0.00659$ \\
Hippocampus & $0.00023$ & $0.00068$ & $0.00065$ & $0.00043$ \\
Whole brain & $0.01181$ & $0.03332$ & $0.02552$ & $0.00601$ \\
Entorhinal cortex & $0.00021$ & $0.00037$ & $0.00032$ & $0.00038$ \\ 
Fusiform & $0.00061$ & $0.00164$ & $0.00196$ & $0.00091$ \\
Middle temporal gyrus & $0.00085$ & $0.00220$ & $0.00263$ & $0.00097$ \\
\bottomrule
\end{tabular}
\end{table*}

\subsection{Irregular time intervals}

As final experiment, we assess generalizability of the proposed method for predicting measurements of irregular visits. In general, standard LSTM networks are designed to handle evenly spaced sequences. We used the same approach in our baseline experiments for AD progression modeling application by disregarding visiting months 3, 6 and 18, and confined the experiments to yearly follow-up in the ADNI data. Now, we employ the available measurements of the 6-th and 18-th visiting months from the TADPOLE dataset and predict biomarker values of half-yearly follow-ups by assuming unavailable visits as missing data. In this experiment, 78\% of the actual data is missing. We apply the same methods to the extended data. Table \ref{table6} details the test modeling performance of the MRI biomarkers for half-yearly predictions using the different DPM methods. As can be seen, our proposed DPM method outperforms all other methods in all categories. More interestingly, considering the corresponding results from Table \ref{table3} for yearly predictions, one can deduce that the modeling performance of the proposed method improves by utilizing the irregular visits. However, the additional time points in the LSTM increases the required time for training and validation to 1,090 seconds and 0.061 seconds, respectively.

As an alternative, one could utilize modified LSTM architectures where the networks learn a number of parameters to encode visiting patterns among longitudinal patient records \citep{baytas2017,neil2016}. However, using such methods not only increase the complexity of the network but also risk learning any time spacing patterns in the data.

\section{Conclusions}

In this paper, a training algorithm was proposed for LSTM networks aiming to improve robustness against missing data, and the robustly trained LSTM network was applied to AD progression modeling using longitudinal measurements of MRI biomarkers. To the best of our knowledge, this is the first time RNNs have been studied and applied to DPM within neurodegenerative disease. Moreover, since RNNs are non-parametric learning methods, the proposed approach can be applied to different time-series data and characteristics than the monotonic behavior that one typically encounters in MRI-based neurodegenerative disease progression modeling. The proposed training method demonstrated better performance than using imputation prior to standard LSTM network training and outperformed an established parametric, regression-based DPM method in terms of both biomarker prediction and subsequent diagnostic classification. This method is also applicable for other types of RNNs such as gated recurrent units (GRUs) \citep{cho2014}. This study highlights the potential of RNNs for modeling the progression of AD using longitudinal measurements, provided that proper care is taken to handle missing values and time intervals.

\section*{Disclosures}

M. Nielsen is shareholder in Biomediq A/S and Cerebriu A/S. A. Pai is shareholder in Cerebriu A/S. The remaining authors report no disclosures.

\section*{Acknowledgments}

This project has received funding from the European Union's Horizon 2020 research and innovation programme under the Marie Skłodowska-Curie grant agreement No 721820. This work uses the TADPOLE data sets (https://tadpole.grand-challenge.org) constructed by the EuroPOND consortium (http://europond.eu) funded by the European Union's Horizon 2020 research and innovation programme under grant agreement No 666992.

Data collection and sharing for this project was funded by the Alzheimer's Disease Neuroimaging Initiative (ADNI) (National Institutes of Health Grant U01 AG024904) and DOD ADNI (Department of Defense award number W81XWH-12-2-0012). ADNI is funded by the National Institute on Aging, the National Institute of Biomedical Imaging and Bioengineering, and through generous contributions from the following: AbbVie, Alzheimer's Association; Alzheimer's Drug Discovery Foundation; Araclon Biotech; BioClinica, Inc.; Biogen; Bristol-Myers Squibb Company; CereSpir, Inc.; Cogstate; Eisai Inc.; Elan Pharmaceuticals, Inc.; Eli Lilly and Company; EuroImmun; F. Hoffmann-La Roche Ltd. and its affiliated company Genentech, Inc.; Fujirebio; GE Healthcare; IXICO Ltd.; Janssen Alzheimer Immunotherapy Research \& Development, LLC.; Johnson \& Johnson Pharmaceutical Research \& Development LLC.; Lumosity; Lundbeck; Merck \& Co., Inc.; Meso Scale Diagnostics, LLC.; NeuroRx Research; Neurotrack Technologies; Novartis Pharmaceuticals Corporation; Pfizer Inc.; Piramal Imaging; Servier; Takeda Pharmaceutical Company; and Transition Therapeutics. The Canadian Institutes of Health Research is providing funds to support ADNI clinical sites in Canada. Private sector contributions are facilitated by the Foundation for the National Institutes of Health (www.fnih.org). The grantee organization is the Northern California Institute for Research and Education, and the study is coordinated by the Alzheimer's Therapeutic Research Institute at the University of Southern California. ADNI data are disseminated by the Laboratory for Neuro Imaging at the University of Southern California.

\bibliographystyle{model2-names}
\bibliography{refs}

\end{document}